\documentclass[letterpaper, 10 pt, conference]{ieeeconf}  

\IEEEoverridecommandlockouts                              

\usepackage{graphics} 
\usepackage{epsfig} 
\usepackage{mathptmx} 
\usepackage{times} 
\usepackage{amsmath} 
\usepackage{amssymb}  

\usepackage{cite}
\usepackage{graphicx}
\usepackage{booktabs}
\usepackage{multirow}

\title{\LARGE \bf
Bag-of-Word-Groups (BoWG): A Robust and Efficient Loop Closure Detection Method Under Perceptual Aliasing}

\author{Xiang Fei$^{1}$, Tina Tian$^{1}$, Howie Choset$^{1}$, Lu Li$^{1,\star}$
\thanks{This work was supported by the Department of Energy (DOE)’s Advanced Research Projects Agency-Energy (ARPA-E), REPAIR Program.}
\thanks{$^{1}$\textit{The authors are with the Biorobotics Lab, the Robotics Institute at Carnegie Mellon University, Pittsburgh, PA 15213, USA}}
\thanks{$^{\star}$Lu Li is the corresponding author. Email: lilu12@andrew.cmu.edu} 
\thanks{© 20XX IEEE.  Personal use of this material is permitted.  Permission from IEEE must be obtained for all other uses, in any current or future media, including reprinting/republishing this material for advertising or promotional purposes, creating new collective works, for resale or redistribution to servers or lists, or reuse of any copyrighted component of this work in other works.}
}

\begin{document}

\maketitle
\thispagestyle{empty}
\pagestyle{empty}

\begin{abstract}
Loop closure is critical in Simultaneous Localization and Mapping (SLAM) systems to reduce accumulative drift and ensure global mapping consistency. However, conventional methods struggle in perceptually aliased environments, such as narrow pipes, due to vector quantization, feature sparsity, and repetitive textures, while existing solutions often incur high computational costs. This paper presents Bag-of-Word-Groups (BoWG), a novel loop closure detection method that achieves superior precision-recall, robustness, and computational efficiency. The core innovation lies in the introduction of word groups, which captures the spatial co-occurrence and proximity of visual words to construct an online dictionary. Additionally, drawing inspiration from probabilistic transition models, we incorporate temporal consistency directly into similarity computation with an adaptive scheme, substantially improving precision-recall performance. The method is further strengthened by a feature distribution analysis module and dedicated post-verification mechanisms. To evaluate the effectiveness of our method, we conduct experiments on both public datasets and a confined-pipe dataset we constructed. Results demonstrate that BoWG surpasses state-of-the-art methods—including both traditional and learning-based approaches—in terms of precision-recall and computational efficiency. Our approach also exhibits excellent scalability, achieving an average processing time of 16 ms per image across 17,565 images in the Bicocca25b dataset. The source code is available at: https://github.com/EdgarFx/BoWG.
\end{abstract}

\section{INTRODUCTION}
Simultaneous Localization and Mapping (SLAM) is a core research area in the field of autonomous robotic systems. It refers that a robot incrementally localizes itself while simultaneously constructing a map of the surroundings. Despite its advancements, SLAM systems often encounter challenges with accumulative drift. This drift significantly constrains the scalability and long-term applicability of SLAM in large and complex operation scenarios.

\begin{figure}[ht]
    \centering
    \includegraphics[width=\linewidth]{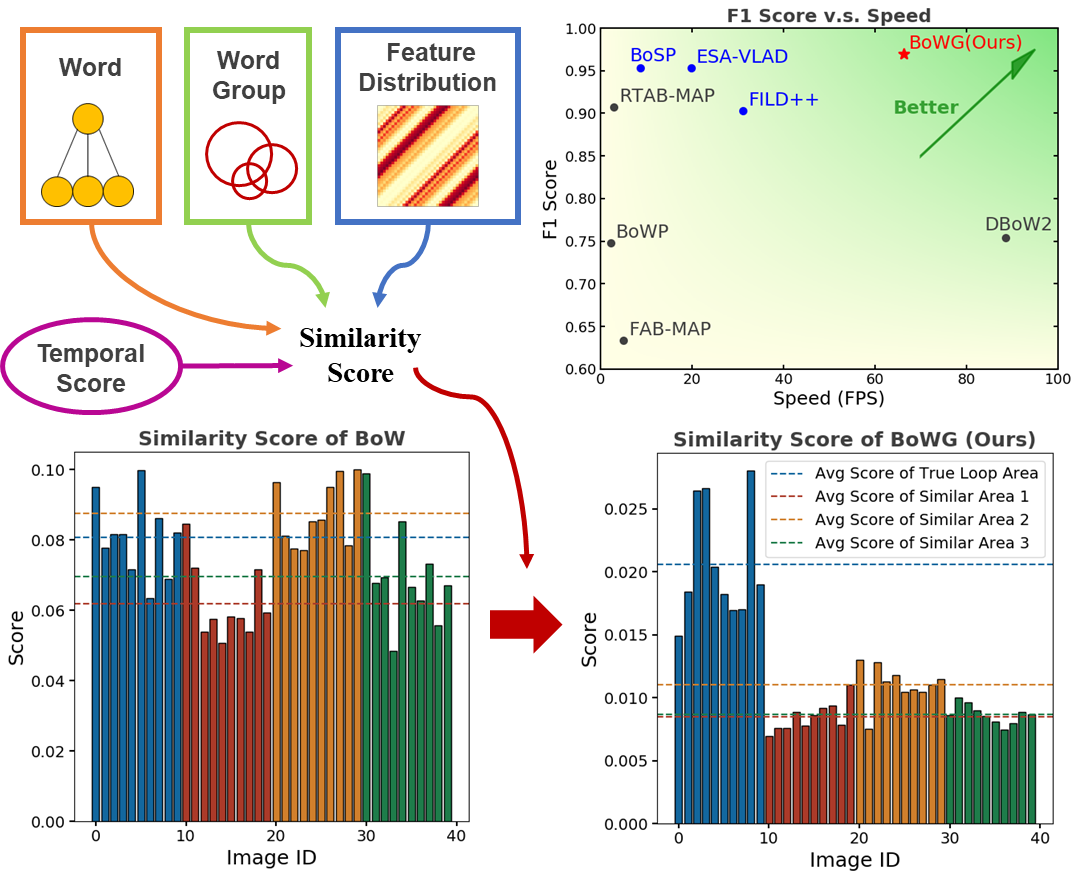}
    \caption{The proposed Bag-of-Word-Groups (BoWG) method enhances loop closure detection by integrating multiple information modalities for similarity computation. The F1 score reflects the precision and recall performance. Learning-based methods (highlighted in blue) were implemented on GPU, while our systems achieves outstanding F1 score with superior execution speed on CPU. In addition, Our system significantly improves the discrimination of the true loop closure area under perceptual aliasing.}
    \label{fig:cover}
\end{figure}

Loop closure is a critical technique in SLAM systems for reducing accumulative drift, thereby improving localization accuracy and mapping consistency \cite{loop}. This approach relies on the robot's ability to accurately recognize previously visited locations and establish constraints between the revisited location (referred to as the loop closure frame) and the current position. The Bag-of-Words (BoW) \cite{tf-idf,dbow2,online_bow} approach is one of the most popular place recognition methods. Although BoW performs effectively in some typical environments, it suffers perceptual aliasing problem \cite{perceptual, perceptual_robust} in challenging environments such as confined pipes \cite{pipe_review}, lava tubes, caves, and mines \cite{dare}, due to its inherent vector quantization issue, compounded by feature scarcity and repetitive textures in such environments. Even in typical environments, BoW demonstrates limited discriminative capability when encountering scenes with high visual similarity.

One approach to address the quantization issue is to directly perform feature matching \cite{direct_hashing, direct_indexing, direct_keyframe} with raw features rather than relying on their quantized representations. While such methods have shown improvements in precision and recall, they are computationally intensive, which makes them impractical for larger-scale applications. In addition, \cite{co_occur} proposed an approach named Bag-of-Word-Pairs (BoWP), which incorporates feature spatial co-occurrence information directly into a multimap dictionary \cite{multimap} to tackle the perceptual aliasing problem. However, this method still constructs a BoW using raw features. In addition, the BoWP method is limited to word pairs with a strict matching criteria that requires computationally expensive SURF features \cite{surf} to ensure sufficient matching pairs and achive advanced performance. Furthermore, although BoWP employs a probabilistic framework \cite{tf-idf} to consider temporal consistency, some of the framework's underlying assumptions (e.g. virtual images) conflict with their design of word pairs.

Aiming to address the aforementioned problems, we propose a novel visual loop closure detection method, termed Bag-of-Word-Groups (BoWG). This method leverages the co-occurrence and proximity information of visual words to create a new dictionary online, enhancing the identification of correct loop closure frame candidates in environments prone to perceptual aliasing. Unlike approaches that rely on raw features or complex descriptors, BoWG is capable of achieving outstanding performance even when using only binary words \cite{brief}. By further integrating the design of direct and inverse index tables, the computational efficiency of BoWG is comparable to that of the well-known and highly efficient BoW model DBoW2 \cite{dbow2}, making it well-suited for large-scale applications. 

Furthermore, while most existing BoW methods only employ temporal consistency information during post-verification \cite{dbow2, temporal, temporal_learning}, we integrate temporal consistency directly into the similarity calculation. Additionally, we observe that the distribution of visual features often offers valuable discriminative information. To leverage this, we design a feature distribution analysis module that integrates this information into the similarity scoring process. Finally, dedicated temporal and geometrical verification modules are developed to further improve the overall performance and reliability of our system. Specifically, the main contributions of this paper are summarized as follows:
\begin{itemize}
    \item A definition of word groups exploiting the co-occurrence and proximity of visual words. This representation enriches the discriminative information of images with similar appearance.
    \item An online word group database is designed and implemented, providing context-specific representation. Through the integration of direct and inverse index tables, our system achieves efficient loop closure detection suitable for large-scale applications.
    \item Temporal consistency and feature distribution information are incorporated directly into the similarity score calculation, complemented by dedicated temporal and geometrical post-verification modules. These additions further improve the system’s precision and recall.
\end{itemize}

\begin{figure*}[htbp]
    \centering
    \includegraphics[width=\linewidth]{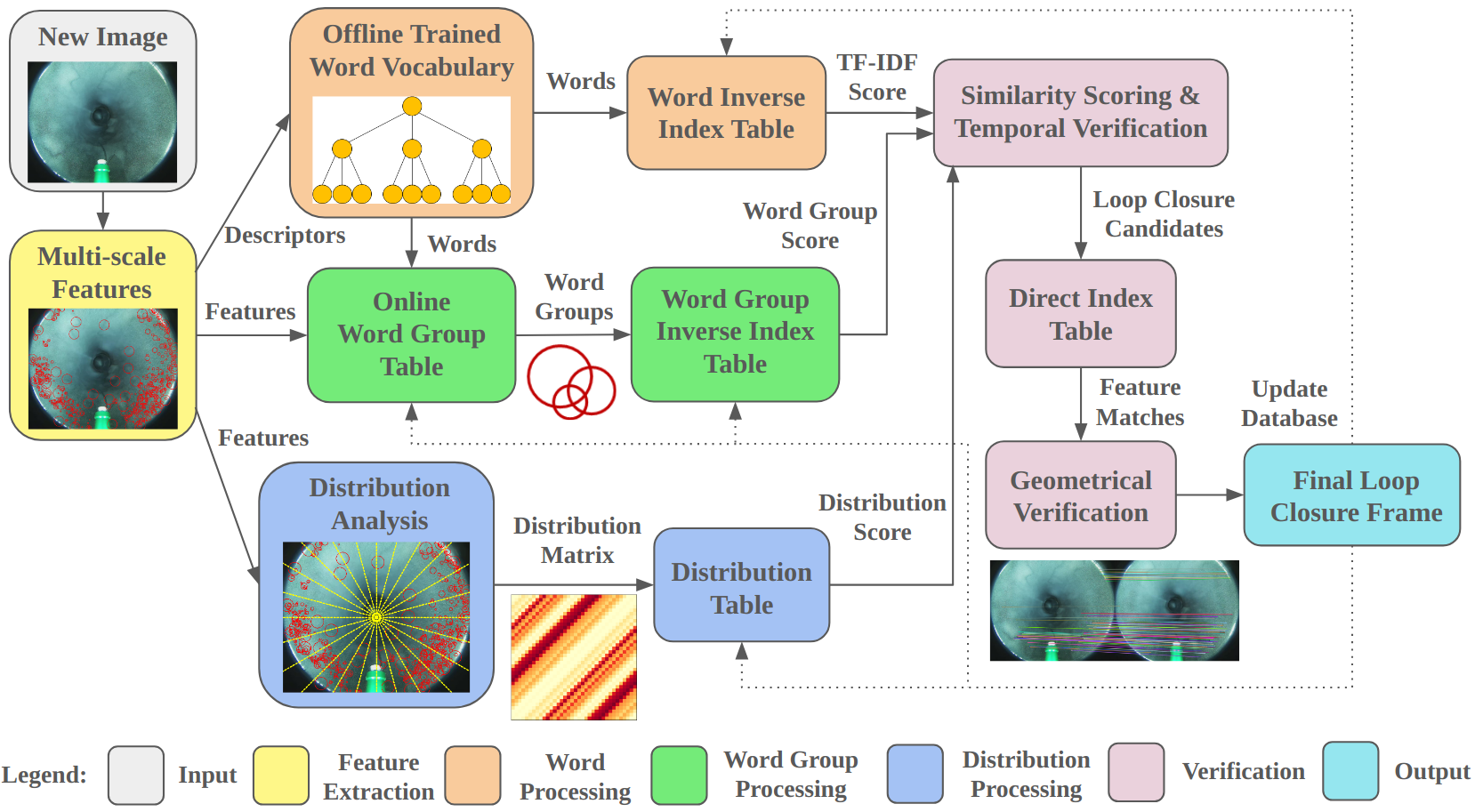}
    \caption{Workflow of the proposed BoWG loop closure detection method. BoWG method consists of five main modules: Feature Extraction, Word Processing, Word Group Processing, Distribution Processing, and Verification.}
    \label{fig:workflow}
\end{figure*}

\section{RELATED WORKS}

\subsection{Bag-of-Words}
The Bag of Words (BoW) technique, originally used in text retrieval\cite{text}, has been widely adopted in place recognition, serving as a crucial component of loop closure detection methods. In the BoW approach, a visual vocabulary is created by clustering the extracted features from a set of training images. Each image is represented as a histogram of visual words (cluster centers), where each bin in the histogram corresponds to the frequency of occurrence of a visual word in the image. To implement loop closure detection, histogram matching is performed between the current and the previous keyframes in the pose graph \cite{direct_keyframe}. 

Bag-of-binary-words (DBoW) \cite{dbow2} is a widely used BoW system that constructs an image database using direct and inverse index files, allowing for efficient image queries, similarity scoring and geometric verification. Its effectiveness has made it a core component of many renowned SLAM systems, such as VINS-Mono \cite{vins} and ORB-SLAM \cite{orbslam}. Subsequent research introduced online BoW methods \cite{online_bow}, which enable real-time database updates without the need for offline training. Other studies have sought to enhance the robustness of BoW models by employing probabilistic frameworks and integrating temporal consistency information \cite{tf-idf, rtab-map}. However, BoW methods suffer perceptual aliasing caused by vector quantization. To address this issue, some approaches propose direct feature matching techniques \cite{direct_hashing, direct_indexing, direct_keyframe}, which bypass the transformation from raw features into quantized words. While these methods improve precision and recall, their high computational cost limits their application in large-scale scenarios.

\subsection{Feature Co-occurrence}
The effectiveness of co-occurring words in addressing the perceptual aliasing problem is well known \cite{co_occur}. FAB-MAP \cite{fab-map} utilizes a Chow–Liu tree to encode co-occurrence information within the observation likelihood of a probabilistic framework. However, while FAB-MAP successfully leverages co-occurrence, it overlooks the spatial proximity of features within an image. This limitation means that two images can be considered similar as long as they share the same set of words, regardless of the spatial arrangement of those words. Moreover, the process of constructing a Chow–Liu tree is computationally intensive and is typically performed offline to satisfy real-time performance constraints.

In later research, \cite{co_occur} considers both the feature co-occurrence and spatial proximity, introducing a multimap dictionary named Bag-of-Word-Pairs (BoWP). While this approach enriches image representation by incorporating visual word pairs, BoWP relies on computationally expensive SURF features \cite{surf} and directly uses raw features as visual words, making it significantly slower compared to efficient BoW methods like DBoW2 \cite{dbow2}. Furthermore, BoWP strictly requires exact word matches between paired words, which largely limits the number of valid matches and reduces its robustness in diverse scenarios. Additionally, BoWP employs a probabilistic framework \cite{tf-idf} to consider temporal consistency. However, this framework involves many assumptions, some of which are incompatible with their design of word pairs.

\section{METHODOLOGY}
\subsection{Bag-of-Word-Groups (BoWG)}

The proposed Bag-of-Word-Groups (BoWG) is a novel loop closure detection method. It analyzes the co-occurrence and proximity information of features for each keyframe, enabling the extraction of more detailed information to improve the differentiation of correct loop closure frames. The workflow is illustrated in Fig. \ref{fig:workflow}. For a new keyframe, we follow the method in \cite{orb} to detect multi-scale features, though our framework supports other detectors as well.

\begin{figure}[htbp]
    \centering
    \includegraphics[width=\linewidth]{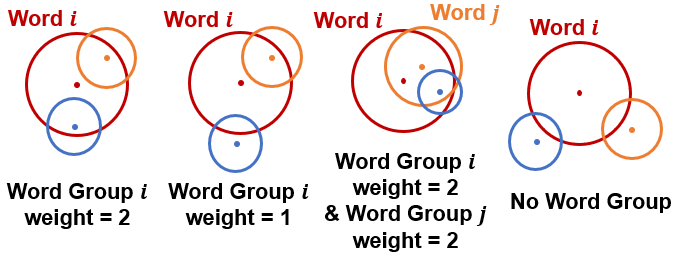}
    \caption{Some examples of word groups.}
    \label{fig:wg_example}
\end{figure}

\textbf{Word Group Definition}:
The word group is designed to capture the co-occurrence and proximity of visual words in an image. For a word $i$ with size $s_t^{(i)}$ in image $t$, all other words whose distance from word $i$ is smaller than $s_t^{(i)}$ and the word $i$ form a word group, and we directly set the word group ID as $i$. The weight of the word group $i$ is defined as the number of the words in the word group minus 1, denoted as $w_{g,t}^{(i)}$. Some examples of word groups are shown in Fig. \ref{fig:wg_example}. 

Inspired by the TF-IDF score \cite{score}, we then refine the word group weight as follows:
\begin{equation}
    \widetilde w_{g,t}^{(i)} = \frac{w_{g,t}^{(i)}}{\sum_i w_{g,t}^{(i)}} \cdot \log \frac{\sum_t\sum_i w_{g,t}^{(i)}}{\sum_t w_{g,t}^{(i)}}
\label{wg_weight}
\end{equation}

\begin{figure}[htbp]
    \centering
    \includegraphics[width=\linewidth]{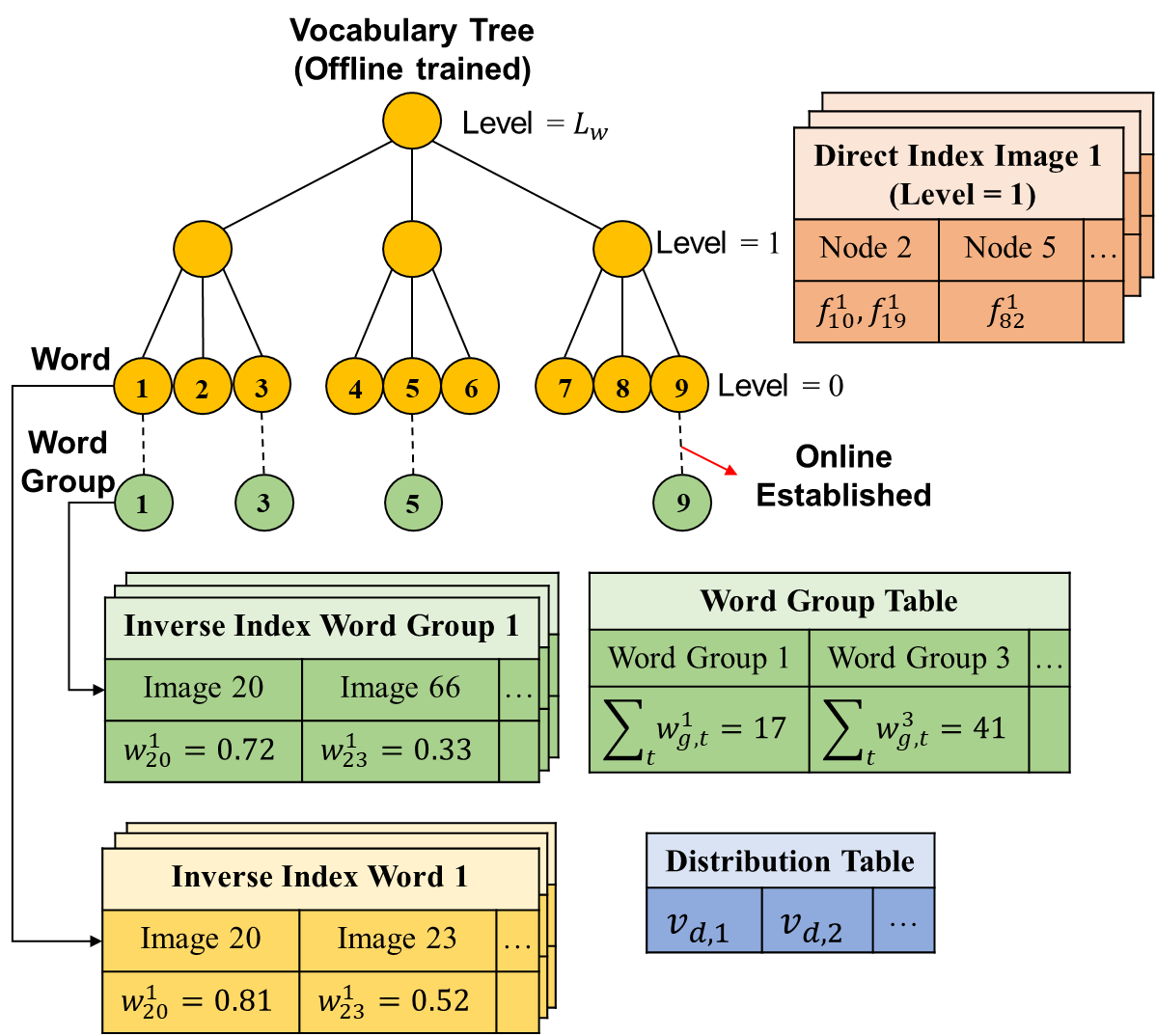}
    \caption{The designed database of our BoWG system.}
    \label{fig:database}
\end{figure}

With our definition, word groups can be regarded as an additional layer of nodes beneath the word nodes in the word vocabulary tree, connecting to parts of the word nodes, as illustrated in Fig. \ref{fig:database}.

\textbf{BoWG Database}:
Our designed BoWG database includes an offline-trained word vocabulary tree, an online word group table, a word inverse index table, a word group inverse index table, a direct index table, and an optional distribution table, as illustrated in Fig. \ref{fig:database}.

\subsubsection{\textbf{Word Vocabulary Tree}}
The word vocabulary tree is used to convert feature descriptors into visual words. It is created offline using a diverse set of features from training images. Following the design in \cite{dbow2}, descriptors are clustered at each level of the tree using $k$-medians clustering with $k$-means++ seeding.

\subsubsection{\textbf{Word Group Table}}
To store word group information, we design a word group table. Since the defined word group is a context-specific concept, the table is built and updated online. Its structure is a map, where the key is the word group ID $i$ and the value stores the cumulative word group weight $i$ across all input images, $\sum_t w_{g,t}^{(i)}$. This design enables efficient computation of the refined word group weight for each word group extracted from a new image.

\subsubsection{\textbf{Inverse Index Table}}
We maintain two inverse index tables: one for words and another for word groups, both utilizing a map as the underlying data structure. The word group inverse index table stores for each word group a list of images in which it appears, and the corresponding weights, facilitating efficient database querying since we can limit comparisons to images that have word groups in common with the query image and quickly access the weights. The word inverse index table follows a similar structure. 

\subsubsection{\textbf{Direct Index Table}}
To enhance the efficiency of geometric verification, we follow the approach in \cite{dbow2} and construct a direct index table. This table stores the features of each image along with their associated nodes at a specific level of the vocabulary tree. This design enables efficient feature matching by restricting comparisons to features belonging to the same node. 

\subsubsection{\textbf{Distribution Table}}
The distribution table is implemented as a vector, where the index corresponds to the image ID and the elements store the distribution vectors of each image. This structure enables efficient retrieval of distribution vectors of previous images and facilitates the computation of the distribution similarity score.

\subsection{Similarity Score}
\subsubsection{\textbf{Word Score}}
In our system, we use TF-IDF weight \cite{score} of words to compute the word score. During the query process, we normalize the word score $s_w(I_t,I_{t_j})$ between the query image $I_t$ and an old image $I_{t_j}$ with the best score we expect to obtain in the sequence for $I_t$, following \cite{normal}:
\begin{equation}
\eta_w(I_t,I_{t_j}) = \frac{s_w(I_t,I_{t_j})}{s_w(I_t,I_{t-1})}
\end{equation}
where we use the word score between image $I_t$ and its closest previous image $I_{t-1}$ as the normalization term.

\subsubsection{\textbf{Word Group Score}}
The word group score is defined as follows:
\begin{equation}
    s_g(I_1,I_2) = \begin{cases}
        1 &\text{,} \sum_i \widetilde w_{g,1}^{(i)}\cdot \widetilde w_{g,2}^{(i)} > 1 \\
        1 - \sqrt{1 - \sum_i \widetilde w_{g,1}^{(i)}\cdot \widetilde w_{g,2}^{(i)}} & \text{,else}
    \end{cases}
\end{equation}
where $\widetilde w_{g,1}^{(i)}$ and $\widetilde w_{g,2}^{(i)}$ are the refined word group weights of the word group with ID $i$ in the two images.

During the query process, we also normalize the word group score as follows:
\begin{equation}
\eta_g(I_t,I_{t_j}) = \frac{s_g(I_t,I_{t_j})}{s_g(I_t,I_{t-1})}
\end{equation}
Then, we can combine the normalized word score and word group score together to obtain the following similarity score:
\begin{equation}
    \eta_{sim} = \lambda_1\eta_w + (1-\lambda_1)\eta_g
\label{similarity}
\end{equation}

\subsubsection{\textbf{Incorporating Temporal Consistency}}
While most existing methods leverage temporal consistency only during post-verification \cite{dbow2, temporal, temporal_learning}, we incorporate it directly into the similarity calculation. Our design is based on the observation that if the previous image $I_{t-1}$ forms a loop closure with $I_{t_j-1}$, it is high likely that the current image $I_t$ also forms a loop closure with $I_{t_j}$. Therefore, the similarity scores of previous queries are relevant to the current score.

In probabilistic loop closure frameworks such as \cite{tf-idf}, only the score of the first previous image is considered based on Markov's Rule. Following this principle, when computing $\eta(I_t, I_{t_j})$, we only consider the influence of $\eta(I_{t-1}, I_{t_j-1})$. To reduce transient errors, we implement an adaptive weighting scheme where the previous score's influence is inversely proportional to the score difference, as follows:
\begin{equation}
    w_{prev} = {w_{max}} / [{1+(\frac{\eta(I_t,I_{t_j})-\eta(I_{t-1},I_{t_j-1})}{\alpha})^2}]
\label{previous_weight}
\end{equation}
where $w_{max}$ is the maximum weight of previous score, and $\alpha$ adjusts the influence of score difference. The similarity score incorporating temporal consistency is as follows:
\begin{equation}
    \eta_{temp} = w_{prev}\cdot \eta(I_{t-1},I_{t_j-1}) + (1-w_{prev}) \cdot \eta(I_t,I_{t_j})
\label{temp_similarity}
\end{equation}

\subsection{Feature Distribution Score}
\begin{figure}[htbp]
    \centering
    \includegraphics[width=\linewidth]{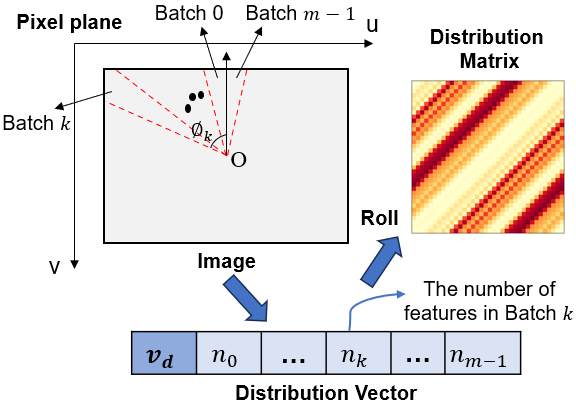}
    \caption{The proposed feature distribution analysis method.}
    \label{fig:distribution}
\end{figure}

When analyzing feature extraction in perceptual aliasing areas such as confined pipes, we observed that images taken from different locations but with similar features often exhibit different feature distributions. This insight led us to enhance the distinguishability of images by comparing their feature distributions. Specifically, we divide the image into $m$ fan-shaped batches, as Fig. \ref{fig:distribution} shown. For batch $k$: $\phi_k = (2\pi / m) \cdot k$, $k$ is from 0 to $m-1$. And then, simply using the number of features in each batch to reflect the distribution of the detected features. Therefore, we can obtain a \textit{Distribution Vector} for each image, whose size is $m$ and the elements are the number of features in each batch:
\begin{equation}
    \textbf{v} = [n_0,n_1,n_2,\cdots,n_{m-1}]^T
\end{equation}
where $\textbf{v}_k = n_k$ is the number of features in batch $k$. The vector is then L2-normalized, denoted as $\mathbf{\hat{v}}$.

However, in environments such as pipes, rotations around the roll axis of the robot are frequent. Since these roll motions can be interpreted as cyclic translations of the vector elements, we define the following $m*m$ \textit{Distribution Matrix}:
\begin{equation}
\mathbf{M} = [m_{ij}], \text{ where } m_{ij} = \hat{\textbf{v}}_{((i+j) \bmod m)}  
\end{equation}
Then, the distribution score between the query image $I_t$ and an old image $I_{t_j}$ is defined as:
\begin{equation}
s_d(I_t,I_{t_j}) = \max\{\|\mathbf{M}_{i}^t - \hat{\mathbf{v}}^{t_j}\|_2 : i = 0, ..., m-1\}
\end{equation}
where $\mathbf{M}_{i}^t$ is the column vector with index $i$ of the distribution matrix of $I_t$, and $\hat{\mathbf{v}}^{t_j}$ is the distribution vector of $I_{t_j}$. This score has some degree of roll-rotation invariance. For cases with large rotation changes, feature detectors with rotation invariance are preferred. We also normalize the score to obtain $\eta_d$. The final similarity score is:
\begin{equation}
\eta_{sim} = \lambda_1\eta_w + \lambda_2\eta_g + (1-\lambda_1-\lambda_2)\eta_d
\label{eq:sim_with_dis}
\end{equation}

The proposed feature distribution analysis is particularly useful in confined spaces such as pipes or scenarios with high-frequency image acquisition. In other environments, this module can be optionally discarded.

\subsection{Loop Closure Determination}


\subsubsection{Islands Matching}
During the query process, if the similarity score $\eta_{sim}(I_t,I_{t_j})$ between the query image $I_t$ and a previous image $I_{t_j}$ is greater than a threshold $\alpha$, we treat $<I_t$,$I_{t_j}>$ as a potential loop closure match. However, images that temporally close to $I_{t_j}$ might also form matches with $I_t$. To avoid competition among such temporally adjacent images when querying the database, we adopt the approach proposed in \cite{dbow2}. Specifically, we group these images into a single entity referred to as an \textit{island} and treat the entire island as a single match. For a time interval $V_{T_i}$ composed of timestamps $t_{n_i}$, $\cdots$, $t_{m_i}$ with small time gap, the island is denoted as $V_{T_i}$. We then compute the island score:
\begin{equation}
    H(I_t,I_{T_i}) = \sum_{j=n_i}^{m_i} \eta_{sim}(I_t,I_{t_j})
\end{equation}
where $\eta_{sim}$ is the proposed similarity score, either \eqref{similarity} or \eqref{eq:sim_with_dis}, depending on whether feature distribution analysis is applied. The island with the highest score is selected as the matching island and proceeds to the temporal verification step. 

\begin{table*}[htbp]
\centering
\caption{Performance Comparison of BoWG with other state-of-the-art algorithms in terms of Recall (\%) at 100\% Precision. A ‘–’ indicates that the corresponding value is not available for a given algorithm and a given dataset.}
\label{tab:recall}
\resizebox{\textwidth}{!}{
\begin{tabular}{c|c|cccccccc}
\toprule
Datasets & \# Images & \begin{tabular}[c]{@{}c@{}} BoWP\cite{co_occur} \end{tabular} & \begin{tabular}[c]{@{}c@{}} FAB-MAP\cite{fab-map} \end{tabular} & \begin{tabular}[c]{@{}c@{}} RTAB-MAP\cite{rtab-map} \end{tabular} & \begin{tabular}[c]{@{}c@{}} DBoW2\cite{dbow2} \end{tabular} & \begin{tabular}[c]{@{}c@{}} FILD++\cite{fild} \end{tabular} & \begin{tabular}[c]{@{}c@{}} BoSP-Topo\cite{bag_superpoints} \end{tabular} & \begin{tabular}[c]{@{}c@{}} ESA-VLAD\cite{esa_vlad} \end{tabular} & \begin{tabular}[c]{@{}c@{}} \textbf{BoWG}\\ \textbf{(Ours)} \end{tabular}
\\ 
\midrule
New College (NC) & 1073 & 59.78$^\dagger$ & 46.73$^\star$ & 82.99 & 60.53 & 82.37$^\dagger$ & 91.00$^\dagger$ & \underline{91.02} & \textbf{94.23} \\
City Centre (CC) & 1237 & 66.80$^\dagger$ & 34.02$^\star$ & 75.04 & 59.63 & 90.01$^\dagger$ & \textbf{92.00}$^\dagger$ & 90.31 & \underline{91.04} \\
Bicocca25b (BI) & 1757 & - & 25.75$^\star$ & \underline{65.67} & 60.07 & - & - & - & \textbf{72.39} \\
\bottomrule
\multicolumn{10}{l}{$^\star$ The precision of FAB-MAP on NC is 98.47\%, on CC is 97.85\%, on BI is 90.79\%.} \\
\multicolumn{10}{l}{$^\dagger$ BoWP, FILD++, and BoSP-Topo used both the left and right images of NC and CC, while other methods (including ours) only used the left images.}\\
\end{tabular}
}
\end{table*}

\begin{table*}[htbp]
\centering
\caption{Average execution time (in seconds) per image for different datasets. The values for BoWP were taken from \cite{co_occur}. The values for other algorithms were obtained through our own testing.}
\resizebox{\textwidth}{!}{
\begin{tabular}{c|c|cccccccc}
\toprule
Datasets & \# Images & \begin{tabular}[c]{@{}c@{}} BoWP\cite{co_occur} \end{tabular} & \begin{tabular}[c]{@{}c@{}} FAB-MAP\cite{fab-map} \end{tabular} & \begin{tabular}[c]{@{}c@{}} RTAB-MAP\cite{rtab-map} \end{tabular} & \begin{tabular}[c]{@{}c@{}} DBoW2\cite{dbow2} \end{tabular} & \begin{tabular}[c]{@{}c@{}} FILD++\cite{fild} \end{tabular} & \begin{tabular}[c]{@{}c@{}} BoSP-Topo\cite{bag_superpoints} \end{tabular} & \begin{tabular}[c]{@{}c@{}} ESA-VLAD\cite{esa_vlad} \end{tabular} & \begin{tabular}[c]{@{}c@{}} \textbf{BoWG}\\ \textbf{(Ours)} \end{tabular}
\\ 
\midrule
New College (NC) & 1073 & 0.4410 & 0.1985 & 0.3384 & \textbf{0.0113} & - & 0.1157$^\star$ & 0.0506$^\star$ & \underline{0.0151} \\
City Centre (CC) & 1237 & 0.3930 & 0.2566 & 0.3514 & \textbf{0.0125} & 0.0321$^\star$ & 0.1122$^\star$ & 0.0501$^\star$ & \underline{0.0186} \\
Bicocca25b (BI) & 1757 & - & 0.4066 & 0.0748 & \textbf{0.0063} & - & - & - & \underline{0.0083} \\
\bottomrule
\multicolumn{10}{l}{
$^\star$ These methods were executed with GPU.} \\
\end{tabular}
\label{tab:time}
}
\end{table*}

\subsubsection{Temporal and Geometrical Verification}
After identifying the best matching island $I_{T'}$, we verify its temporal consistency as \cite{dbow2}. The match $<I_t$,$I_{T'}>$ must be consistent with $k$ previous matches $<I_{t-\Delta t}$,$I_{T_1}>$, $\cdots$, $<I_{t-k\Delta t}$,$I_{T_k}>$, such that the intervals $T_j$ and $T_{j+1}$ are close to overlap. If the island $I_{T'}$ satisfies this temporal constraint, we retain only the match $<I_t$,$I_{t'}>$, where $t' \in T'$ is chosen to maximize the similarity score $\eta_{sim}$. 

The obtained matching image $I_{t'}$ is then passed to the geometrical verification module for further validation. This process involves identifying a fundamental matrix between $I_t$ and $I_{t'}$ using Random Sample Consensus (RANSAC). The most time-consuming aspect of this validation is computing the correspondences, which requires comparing the local features of the query image with those of the matched image. However, by leveraging the direct index table, we can significantly accelerate the execution while maintaining comparable recall and precision to exhaustive search methods.

\section{EXPERIMENTS}

In this section, we evaluate the performance of our proposed BoWG method by comparing it against state-of-the-art methods on both public datasets and real-world pipe environments with significant perceptual aliasing. 

\textbf{Word Vocabulary Training}. Our vocabulary tree is constructed with $k_w=10$ branches and $L_w=6$ depth levels, resulting in a total of one million visual words. The tree is trained using 20,000 images from the independent dataset \textit{Bovisa 2008-09-01} \cite{rawseeds}, which features both outdoor and indoor images captured in static scenarios.

\subsection{Experiments on Public Datasets}
In this subsection, we evaluate our system's performance against seven leading algorithms in the field. The comparison includes traditional methods such as BoWP \cite{co_occur}, FAB-MAP \cite{fab-map}, RTAB-MAP \cite{rtab-map}, and the bag of binary words approach (DBoW2) \cite{dbow2}. Additionally, we also compare against contemporary learning-based approaches: FILD++ \cite{fild}, which utilizes deep features and proximity graphs; the bag-of-superpoints method with graph verification (BoSP-Topo) \cite{bag_superpoints}; and ESA-VLAD \cite{esa_vlad}, which combines second-order attention mechanisms with NetVLAD \cite{netvlad}. In addition, the execution time of our method is then analyzed, verifying its efficiency and scalability. We also conduct ablation studies to show the effectiveness of the modules in our system. 

We evaluated our method on three widely-used datasets: \textit{New College} \cite{new_college}, \textit{City Centre} \cite{rtab-map}, and \textit{Bicocca25b} \cite{rawseeds}. New College contains repetitive architectural features like identical archways, uniform walls and bushes. City Centre, collected from urban roads, presents challenges from dynamic objects and environmental variations. Bicocca25b covers office buildings with repetitive indoor scenes such as corridors and libraries. In the experiments on these datasets, the feature distribution analysis is disabled.


\textbf{Recall \& Precision Comparison.} The performance of the system is evaluated in terms of precision-recall metrics. The setting of this experiment is similar to that of \cite{rtab-map}. The only difference is that for New College and City Centre datasets, we only use the left camera image rather than merging the left and right images. For the Bicocca25b dataset, we follow the setting in \cite{rtab-map}, where some images were removed to achieve an approximate image rate of 1 Hz, keeping 1,757 of the original 52,695 images. 
 
Table \ref{tab:recall} compares the recall of our BoWG method with other algorithms at 100\% precision on these datasets. Results for BoWP \cite{co_occur}, FILD++ \cite{fild}, BoSP-Topo \cite{bag_superpoints}, and ESA-VLAD \cite{esa_vlad} are from their papers, while others are based on their open-source implementations and our experiments. Missing values are marked as ‘–’. The experimental results demonstrate that our method achieves competitive or superior recall compared to learning-based approaches and substantially outperforms traditional methods, despite using only monocular images, while some other methods use stereo imagery to obtain the results shown.

\begin{figure}[htbp]
    \centering
    \includegraphics[width=\linewidth]{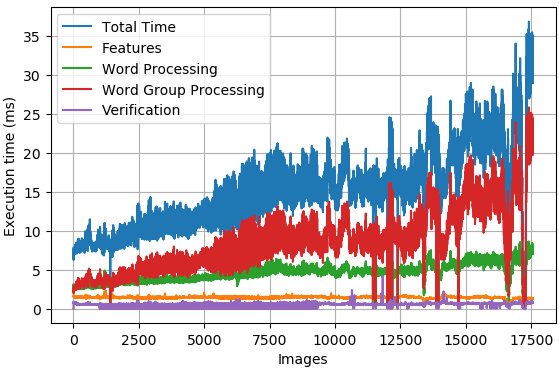}
    \caption{Execution time in Bicocca25b with 17565 images.}
    \label{fig:time}
\end{figure}

\textbf{Execution Time.} Table \ref{tab:time} compares the average per-image processing time across datasets. Our experiments ran on an Intel Core i9 @2.2 GHz CPU, while learning-based methods used GPUs \cite{fild, bag_superpoints, esa_vlad}. The proposed BoWG achieves efficiency closest to DBoW2 \cite{dbow2} and significantly outperforms other methods. To evaluate scalability, we tested on Bicocca25b at 10 Hz (17,565 images), with results shown in Fig. \ref{fig:time}. Despite word group processing being the most intensive component, it remains efficient with only a slow increase in time as the database grows. BoWG processes queries in 16 ms average (37 ms maximum) on the 17,565 images dataset, while FAB-MAP requires 407 ms on just 1,757 images — our method is 25 times faster on a 10 times larger database. This demonstrates BoWG's suitability for large-scale applications.

\begin{figure}[htbp]
    \centering
    \includegraphics[width=0.9\linewidth]{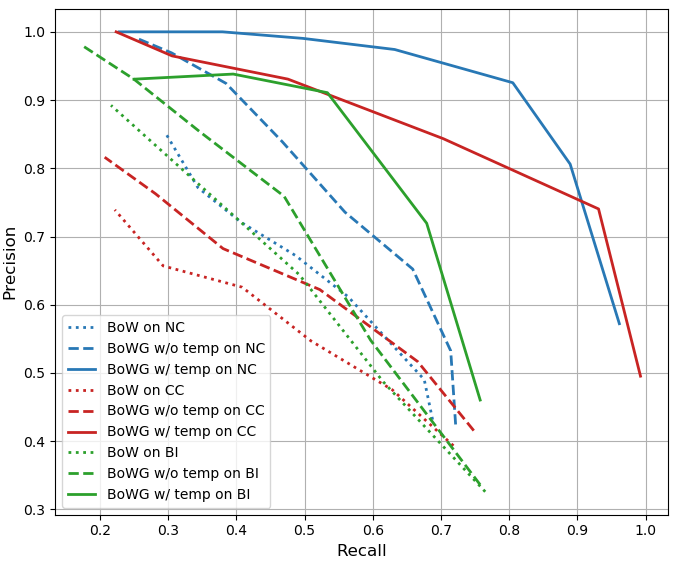}
    \caption{Precision–recall curve of pure BoW, and BoWG with/without the proposed adaptive temporal score on New College, City Centre, and Bicocca25b datasets with no temporal and geometrical check.}
    \label{fig:word_group_res}
\end{figure}

\textbf{Word Groups and Temporal Score.} To assess the impact of the proposed word group and adaptive temporal score, we tested our system on the three datasets without post-verification. As shown in Fig. \ref{fig:word_group_res}, integrating the word group information and the adaptive temporal score largely improves precision-recall performance. We will further analyze the effectiveness of word group later in a real-world pipe environment, where perceptual aliasing presents greater challenges.
\begin{figure}[htbp]
    \centering
    \includegraphics[width=\linewidth]{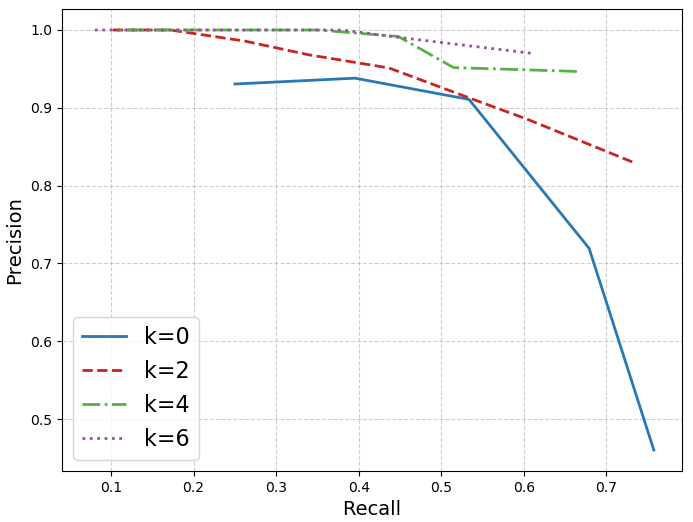}
    \caption{Precision–recall curve in Bicocca25b with no geometrical check, for several values of temporally consistent matches k.}
    \label{fig:temporal}
\end{figure}

\begin{table}[htbp]
\centering
\caption{Performance of Different Approaches to Obtain Correspondence on New College}
\begin{tabular}{|l|c|cccc|}
\hline
\multicolumn{1}{|c|}{\multirow{2}{*}{Technique}} & \multirow{2}{*}{Recall (\%)}& \multicolumn{4}{c|}{Execution Time (ms / image)}                                               \\ \cline{3-6} 
\multicolumn{1}{|c|}{}                           &                              & \multicolumn{1}{c|}{Median} & \multicolumn{1}{c|}{Mean}  & \multicolumn{1}{c|}{Std}   & Max    \\ \hline
DI at Level 3                                              & 79.09                        & \multicolumn{1}{c|}{0.216}  & \multicolumn{1}{c|}{0.935} & \multicolumn{1}{c|}{1.058} & 2.516  \\ \hline
DI at Level 4                                             & 92.79                        & \multicolumn{1}{c|}{0.235}  & \multicolumn{1}{c|}{0.683} & \multicolumn{1}{c|}{0.922} & 2.489  \\ \hline
DI at Level 5                                            & 94.23                        & \multicolumn{1}{c|}{0.854}  & \multicolumn{1}{c|}{0.664} & \multicolumn{1}{c|}{0.615} & 3.944  \\ \hline
Exhaustive                                       & 94.71                        & \multicolumn{1}{c|}{7.093}  & \multicolumn{1}{c|}{4.649} & \multicolumn{1}{c|}{3.577} & 10.909 \\ \hline
\end{tabular}
\label{tab:geometric}
\end{table}

\textbf{Temporal and Geometrical Verification.} To evaluate the effectiveness of our temporal consistency checking, we tested our system on Bicocca25b while varying the required number of temporally consistent matches \textit{k}. Fig. \ref{fig:temporal} shows that temporal verification ($k>0$) clearly improves precision while maintaining high recall compared to no verification case ($k=0$). For geometrical verification, we leveraged the direct table from \cite{dbow2} to reduce descriptor comparisons during point correspondence computation. As shown in Table \ref{tab:geometric}, while all techniques achieve 100\% precision, using the direct index (DI) table maintains high recall with substantially reduced computation time compared to exhaustive search.


\subsection{Experiments on More Challenging Pipe Environments}

\begin{figure}[htbp]
    \centering
    \includegraphics[width=\linewidth]{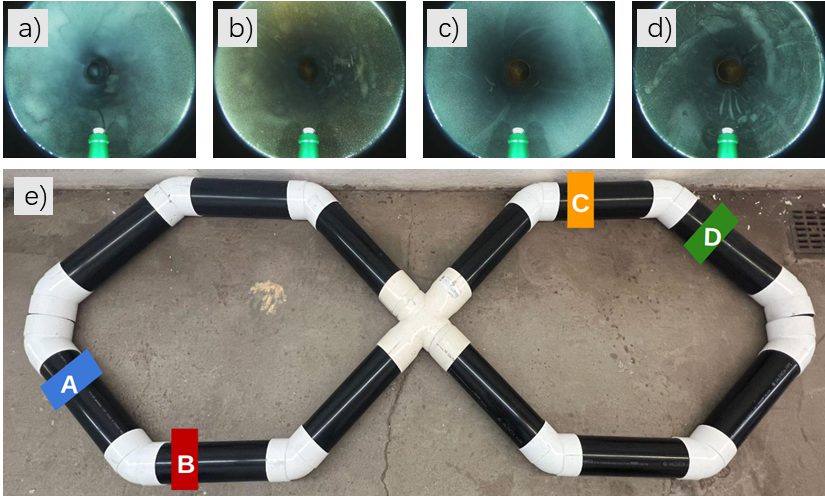}
    \caption{(a)-(d) Example images in area A-D (e) The 6'' pipe test site.}
    \label{fig:test_site}
\end{figure}

\begin{figure}[htbp]
    \centering
    \includegraphics[width=\linewidth]{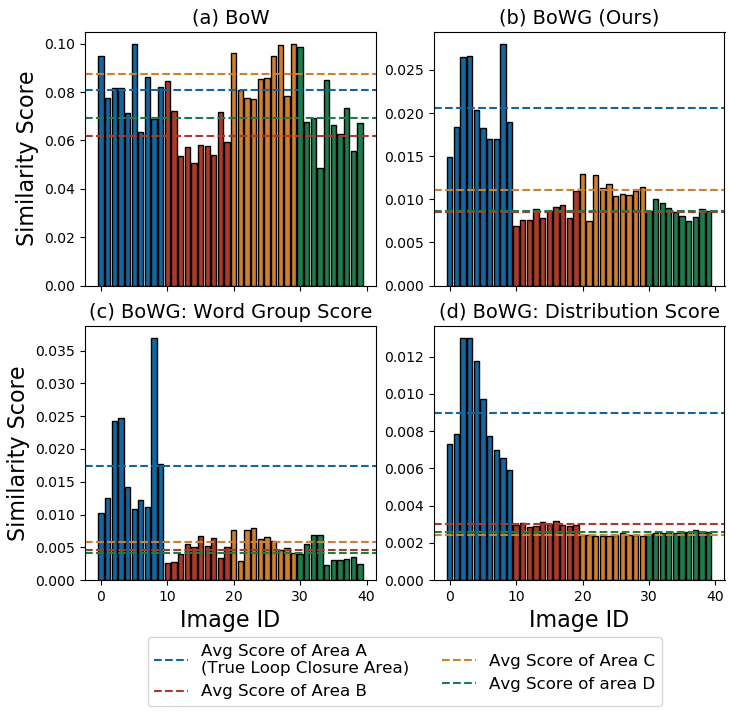}
    \caption{Similarity scores obtained by different methods when area A is the true loop closure area. The first row shows the scores obtained by (a) the pure BoW and (b) our proposed BoWG method. The second row displays the (c) word group score and (d) distribution score of our method.}
    \label{fig:a_scores}
\end{figure}


To test loop closure detection under severe perceptual aliasing, we constructed a test environment using 6'' PVC pipes (Fig. \ref{fig:test_site}). The narrow pipe environment, characterized by feature scarcity and textural repetitiveness, provides an ideal setting for this evaluation. We analyzed similarity scores between a query image from area A (true loop closure area) and 40 reference images evenly distributed across four distinct areas (A, B, C, D). In this test, Good Features to Track (GFTT) detector \cite{gftt} is used.

\textbf{Test Results.}
Fig. \ref{fig:a_scores} compares similarity scores for 40 frames against a newly collected frame using different methods and BoWG modules. As shown in Fig. \ref{fig:a_scores} (a), the pure BoW method assigns a higher average similarity score to area C than area A, with some frames from areas B and D also exceeding those from A, highlighting the severity of perceptual aliasing.

In contrast, BoWG consistently assigns higher similarity scores to area A, as shown in Fig. \ref{fig:a_scores} (b). The second row further decomposes the contributions of different modules. As illustrated in Fig. \ref{fig:a_scores} (c), the word group significantly enhances the discrimination of true loop closure area A, with its average similarity score markedly surpassing those of other areas. Moreover, all frames within area A maintain consistently higher scores. Notably, the highest word group score in area A far exceeds those in other areas, facilitating the selection of an appropriate similarity threshold and improving precision in practice. This is particularly crucial in SLAM loop closure detection, where incorrect detection and relocalization pose a greater risk than accumulative drift.

Furthermore, the proposed feature distribution scores can also effectively help us distinguish the correct loop closure area A from other areas, as shown in Fig. \ref{fig:a_scores} (d). In particular, we can see that the distribution score of area C is significantly lower than that of area A. This shows that even though area C and area A have many common features, these features have different aggregation characteristics in these two areas, verifying our motivation to leverage feature distribution information.


\section{CONCLUSION AND DISCUSSION}
This paper proposes Bag-of-Word-Groups (BoWG), a robust and efficient method for loop closure detection under perceptual aliasing. Our system demonstrates superior performance over state-of-the-art traditional and learning-based methods, in terms of precision-recall metrics and computational efficiency. However, several aspects of our method could be further improved. First, compared to the pure BoW approach, BoWG introduces additional hyperparameters that may require tuning to achieve optimal performance across environments that differ substantially, although our method can achieve significant improvements over pure BoW even with non-optimal parameter settings in practice. Additionally, our proposed feature distribution analysis has limited applicable scenarios, extending its applicability to a broader range of scenarios is a promising direction for future research.

\section*{ACKNOWLEDGMENT}

This work was supported by the Department of Energy (DOE)’s Advanced Research Projects Agency-Energy (ARPA-E), REPAIR Program; Carnegie Mellon University Robotics Institute Summer Scholar (RISS) program; the Biorobotics Lab; Shenzhen Institute of Artificial Intelligence and Robotics for Society (AIRS); and the Chinese University of Hong Kong, Shenzhen (CUHK, Shenzhen).

\bibliographystyle{abbrv}
\bibliography{references}

\end{document}